\documentclass[10pt,twocolumn,letterpaper]{article}

\usepackage[algorithms]{wacv}      %
\usepackage[accsupp]{axessibility} %

\usepackage{graphicx}
\usepackage{amsmath}
\usepackage{amssymb}
\usepackage{booktabs}
\usepackage{algorithm}
\usepackage{algpseudocode}
\usepackage{times}
\usepackage{relsize}
\usepackage{epsfig}
\usepackage[export]{adjustbox}

\usepackage{siunitx}
\usepackage{soul}
\usepackage{xcolor}

\usepackage{pifont}%
\newcommand{\cmark}{\ding{51}}%
\newcommand{\xmark}{\ding{55}}%

\usepackage{tikz}
\usetikzlibrary{automata, positioning}

\usepackage[accsupp]{axessibility} %
\usepackage{cite}
\usepackage{url}
\usepackage{amsmath,amssymb,amsfonts}
\usepackage{epsfig}

\usepackage[]{graphicx}

\usepackage{textcomp}
\def\BibTeX{{\rm B\kern-.05em{\sc i\kern-.025em b}\kern-.08em
    T\kern-.1667em\lower.7ex\hbox{E}\kern-.125emX}}
\usepackage{fancyhdr}
\def\BibTeX{{\rm B\kern-.05em{\sc i\kern-.025em b}\kern-.08em
    T\kern-.1667em\lower.7ex\hbox{E}\kern-.125emX}}
\usepackage{float}
\usepackage{multirow}
\usepackage{enumitem}
\setlist{leftmargin=3.5mm}
\usepackage{mathtools}
\usepackage{bm}
\usepackage{mathtools,amssymb,lipsum, nccmath}

\usepackage{cuted}
\usepackage{mathptmx} %
\usepackage{times} %
\usepackage{booktabs} %
\usepackage{tabularx}
\usepackage{colortbl}

\usepackage{algpseudocode}%
\makeatletter
\def\BState{\State\hskip-\ALG@thistlm}
\makeatother
\usepackage{textcomp}
\usepackage{array}

\usepackage{epsfig} %
\makeatletter

\usepackage[pagebackref,breaklinks,colorlinks]{hyperref}

\usepackage[capitalize]{cleveref}
\crefname{section}{Sec.}{Secs.}
\Crefname{section}{Section}{Sections}
\Crefname{table}{Table}{Tables}
\crefname{table}{Tab.}{Tabs.}

\def\wacvPaperID{03} %
\def\confName{WACV}
\def\confYear{2024}

\begin{document}

\title{Co-Speech Gesture Detection through Multi-Phase Sequence Labeling}

\author{
Esam Ghaleb$^{1}$ \; Ilya Burenko$^{2,3}$\; Marlou Rasenberg$^{4,6}$\; Wim Pouw$^{5}$\; Peter Uhrig$^{2,3}$\;\\
Judith Holler$^{5,6}$\; Ivan Toni$^{5}$\; Asl\i~Özy\"{u}rek$^{5,6}$ and Raquel Fern\'{a}ndez$^{1}$\\
$^{1}$University of Amsterdam\;
$^{2}$ScaDS.AI Dresden/Leipzig\; 
$^{3}$TU Dresden\; 
$^{4}$Meertens Institute\\
$^{5}$Radboud University\; 
$^{6}$Max Planck Institute for Psycholinguistics\\
{\tt\small e.ghaleb@uva.nl \;\;raquel.fernandez@uva.nl}
}
\maketitle
\thispagestyle{empty}
\begin{abstract}
Gestures are integral components of face-to-face communication. They unfold over time, often following predictable movement phases of preparation, stroke, and retraction. Yet, the prevalent approach to automatic gesture detection treats the problem as binary classification, classifying a segment as either containing a gesture or not, thus failing to capture its inherently sequential and contextual nature. To address this, we introduce a novel framework that reframes the task as a multi-phase sequence labeling problem rather than binary classification. Our model processes sequences of skeletal movements over time windows,  uses Transformer encoders to learn contextual embeddings, and leverages Conditional Random Fields to perform sequence labeling. We evaluate our proposal on a large dataset of diverse co-speech gestures in task-oriented face-to-face dialogues. The results consistently demonstrate that our method significantly outperforms strong baseline models in detecting gesture strokes. Furthermore, applying Transformer encoders to learn contextual embeddings from movement sequences substantially improves gesture unit detection. These results highlight our framework's capacity to capture the fine-grained dynamics of co-speech gesture phases, paving the way for more nuanced and accurate gesture detection and analysis.

\end{abstract}

\section{Introduction}
\label{sect:introduction}
\begin{figure}[t!]
\centering
\includegraphics[width=\columnwidth]{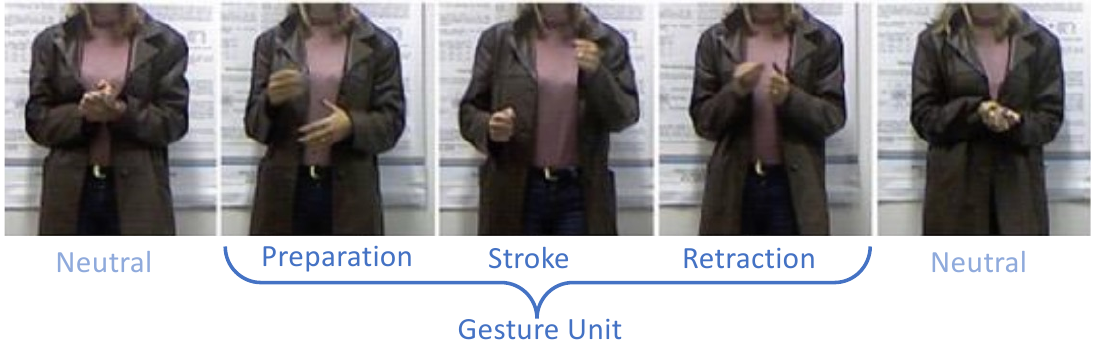}
\caption{A gesture unit consists of sequential gestural phases. Figure adapted from Sanchez \etal \cite{sanchez2022gesture}.}
\label{fig:gestural_phases}
\vspace*{-10pt}
\end{figure}

Gestures are inherent ingredients of face-to-face communication %
that serve many functions, such as illustrating objects or actions, emphasizing verbal expression, or indicating direction~\cite{mcneill1992hand}. Gesture analysis is a key research area in human-computer interaction, sign language recognition, and behavior analysis, where sensory data gathered through  motion capture or glove-based sensors can be leveraged \cite{guo2021human}. Despite the accuracy of these technologies, wearing sensors may be considered intrusive by users. Consequently, passive sensors such as RGB or depth cameras have been widely adopted for gesture analysis. Using  data gathered through such passive sensors, vision-based gesture detection and recognition models are currently the most dominant in the field \cite{molchanov2016online, zhu2018continuous, kopuklu2019real, benitez2021ipn}. 
Recent studies, however, have two main limitations. 
First, widespread gesture detection methods, such as classification techniques, often apply a \textit{binary approach}, e.g., classifying each video frame or segment as either gestural or non-gestural \cite{kopuklu2019real, molchanov2016online, zhu2018continuous}. They therefore do not exploit the fact that gestures consist of different phases, typically including preparation, stroke, and retraction, as illustrated in Figure~\ref{fig:gestural_phases}.
Second, the predominant body of literature focuses on analyzing a limited range of \textit{isolated gestures in controlled conditions}, often depicting discrete actions or objects \cite{benitez2021ipn, molchanov2016online}.
However, in actual communicative situations, gestures are used dynamically as part of face-to-face conversation.  
This research aims to address these two limitations by emphasizing gestures' complex and sequential nature, focusing on detecting co-speech gestures in naturalistic, conversational data.

Gestures are not abrupt events; instead, they unfold over time, often following a predictable pattern of phases. As characterized by Kendon \cite{kendon2004gesture}, when a person gestures, certain body parts undertake a ``movement excursion.'' This excursion, depicted in Figure \ref{fig:gestural_phases}, begins when the gesturing body parts depart from a rest position and ends when they return to it. This process is considered a \textit{gesture unit}, composed of well-defined movement patterns. Within this unit, we can distinguish one or more phases \cite{kendon2004gesture}. 
The preparation phase leads up to the stroke---the most meaning-bearing part of the gesture---, while the retraction phase follows it, signaling the completion of the gesture unit. 

We argue that a fine-grained labeling approach is needed to exploit the patterned structure of gestures, an alternative to binary classification. 
This perspective is inspired by sequence labeling strategies employed in %
Natural Language Processing\cite{jurafsky2020sequence}, where we may need to identify the span of a given subsequence of interest (e.g., a named entity such as \textit{``Palace of Westminster''}) within a text\cite{yan2019tener}.
Similarly, we propose to conceptualize gesture detection as  %
a segmentation task requiring sequential modeling of gesture phases. This can be efficiently achieved using a \textit{multi-phase sequence labeling} approach that tracks the lifecycle of each gesture, beginning from an initial neutral position, transitioning into preparation, followed by the execution of the stroke, retraction, and finally returning to the neutral state of rest once again. 

To the best of our knowledge, such a multi-phase sequence labeling approach
has not been employed in previous work on gesture detection. Its introduction for gesture detection represents a change in how we operationalize and model gestures computationally. %
Moreover, sequence-based labeling is particularly suited for powerful models such as Transformers \cite{vaswani2017attention} and learning algorithms such as Conditional Random Fields (CRFs) \cite{lafferty2001conditional}. 
To operationalize gesture detection as a sequence labeling task, we present a framework that utilizes a series of sliding time windows, each representing skeletal movements as spatio-temporal graphs. Spatio-Temporal Graph Convolutional Networks \cite{yan2018spatial, jiang2021skeleton, ghaleb2021skeleton} are used to embed the bodily movements of each time window, capturing the spatio-temporal dynamics of the bodily movements in a short window. Subsequently, embeddings are fed into Transformer encoders \cite{vaswani2017attention}, which encode the longer temporal dependencies across the sequence. We use CRFs for structured prediction, thereby exploiting the labels' sequential dependencies in the gestural unit. 
In short, these are our key contributions and findings: 
\begin{itemize}[topsep=1pt,itemsep=0ex,partopsep=1ex,parsep=0ex]
    \item In this paper, we present the first study on deep learning for co-speech gesture detection, a research domain that has yet to attract attention in computer vision. Beyond this, we introduce a novel framework, operationalizing gesture detection as a multi-phase sequence labeling task rather than a binary classification problem as detailed in Section \ref{sect:proposed_model}, using Transformers and CRFs.
    \item Our study presents in-depth experimental results and evaluations conducted on a rich dataset of co-speech gestures from 38 individuals in face-to-face communication. This dataset, comprising 16 hours of recorded naturalistic dialogue, goes beyond conventional categorical gestures, as introduced in Section \ref{sect:dataset}.
    
    \item Our multi-phase sequence labeling approach to
    gesture detection consistently outperforms traditional binary and multiclass classification methods, as evidenced by the results presented in Section \ref{sect:results} and analyzed in Section \ref{sect:evaluations}. 
    All data and corresponding code are available on GitHub at {\small \url{https://github.com/EsamGhaleb/Multi-Phase-Gesture-Detection}}.\footnote{Code and computational experiments were contributed equally by the first two authors.}

\end{itemize}
\section{Related Work}\label{sect:related_work}

\paragraph{Data Modeling in Gesture Analysis}
\label{subsect:representations}

Current gesture analysis approaches frequently employ deep learning models for feature representation of raw RGB or depth data, as well as for gesture detection and classification. 
A pioneering approach by Molchanov et al. \cite{molchanov2016online} %
uses a 3D-Convolutional Neural Network (CNN) on depth and RGB data and incorporates Connectionist Temporal Classification (CTC) to predict an ``in progress'' gesture from video segments. Typically, the class content of these segments features a \textit{``silent gesture that depicts an action or an object''}.
Köpüklü et al. \cite{kopuklu2019real} take a different path, using RGB video data first to decide if a segment contains a gesture (gesture detection). If confirmed, a 3D-CNN is subsequently employed to classify the gesture content within the segment (gesture recognition).

Recent advancements in pose and keypoint estimation techniques have also enabled precise measurement and tracking of human body positions and 2D and 3D locations of body joints \cite{sengupta2020mm}. This provides a feasible alternative to sensor-based point estimation. As such, keypoint-based models have been employed for gesture analysis in recent studies \cite{liu20203d}, showing notable success in domains such as sign language recognition \cite{jiang2021skeleton}.
Jiang et al. \cite{jiang2021skeleton}, e.g.,  proposed a multi-channel approach for this domain, incorporating RGB, depth, and keypoint-based models. The study, notably, deployed Graph Convolutional Networks (GCNs) with keypoint-based models displaying the highest performance, resulting in the highest recognition rates. Inspired by Jiang et al., our work utilizes GCNs on spatio-temporal graphs constructed using sequences of estimated key points.

\paragraph{Gesture Detection}
Gesture detection approaches predominantly fall into two primary categories. The first family of approaches treats gesture detection as a preprocessing phase before gesture recognition, e.g., by leveraging binary classification, \cite{kopuklu2019real,kong2014towards}, and gesture boundary detection \cite{zhu2018continuous}. For example, Zhu et al.~\cite{zhu2018continuous} proposed 3D-Residual CNNs and Recurrent Neural Networks for gesture boundary detection. 
Binary approaches are also common in sign language recognition. %
Kong and Ranganath's work \cite{kong2014towards} exemplifies this, proposing a model that segments sign language sequences into two categories: SIGN and Movement Epenthesis, facilitated by Bayesian Networks. The segmented SIGN subsegments are subsequently amalgamated, and a combination of CRFs and Support Vector Machines are employed. It is important to note that while this methodology deploys sequence labeling, the labeling scheme is binary.

The second set of approaches integrates gesture detection within the gesture recognition task \cite{song2012continuous, molchanov2016online, morency2007latent}. This procedure is made possible due to the fixed target number of gestures, with non-gestural movements designated as a ``background'' class. For instance, Molchanov et al. \cite{molchanov2016online} applied CTC to categorize gesture classes, including the ``no gesture'' class. Another noteworthy example is the early work of Morency et al. \cite{morency2007latent} in 2007, which utilized Latent-Dynamic(LD)-CRFs to learn intra- and inter-class dynamics for gesture recognition. The study proposed sequential labeling for the recognition of fixed-sized gestures and ``background'' movements with LD-CRFs. However, despite applying sequential labeling in gesture analysis, this work diverges from ours in several ways. First, its emphasis was primarily on gesture recognition; secondly, it did not take into account any sequential phases within gestural units, instead focusing on learning intra-class dynamics over gesture units' dynamics; and finally, it used relatively simple hand-crafted features, as was the norm at the time.

\begin{figure}
    \centering
    \includegraphics[width=0.75\columnwidth]{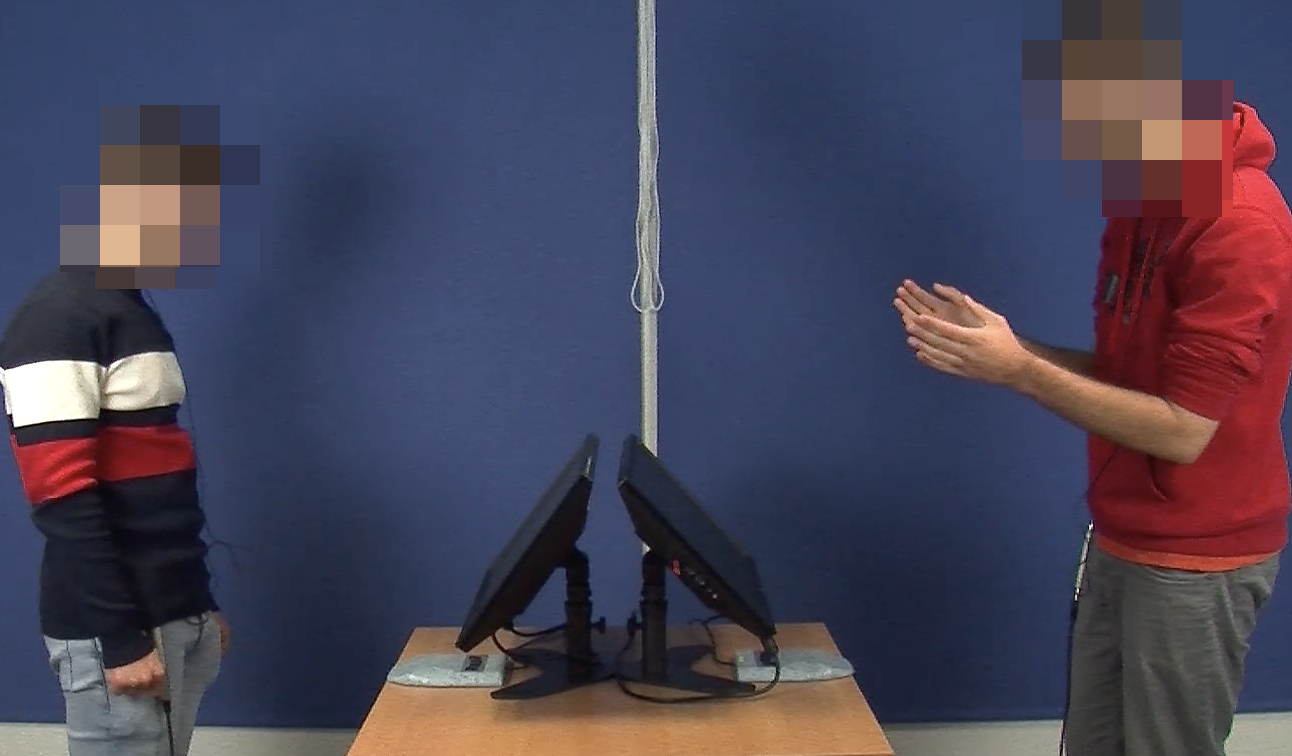}
    \caption{Data collection setup: Two participants play a referential game, freely communicating using speech and gestures.}
    \label{fig:cabb_game_picture}
\end{figure}

\section{Data and Preprocessing}

\subsection{Dataset}
\label{sect:dataset}
Several gesture datasets are available, typically comprising short video clips illustrating silent gestures associated with specific actions or objects \cite{benitez2021ipn, molchanov2016online}. The present study goes beyond identifying %
a limited set of gestures and instead addresses the more challenging task of co-speech gesture detection in naturalistic communication. In particular, we leverage a dataset created by Rasenberg et al.~\cite{rasenberg2022primacy}, which contains 19 face-to-face task-oriented dialogues by 38 different subjects over 16 hours of footage. In this dataset, speakers participate in a referential game, where one participant %
describes a novel object (\ie, a 3D object without a conventional name) while the other participant %
tries to find it among $16$ candidates displayed on a screen, using any means of communication they may want to employ. %
This task design elicited not only speech but also spontaneous gestures, particularly iconic or referential co-speech gestures, as the participants conceptualized the novel object under discussion \cite{rasenberg2022primacy, eijk2022cabb}.
Figure \ref{fig:cabb_game_picture} shows the setup of the data collection, where videos were recorded from different angles for each participant.
We use recordings from cameras positioned to the side, which result in semi-frontal views of each speaker.

Each gesture stroke was manually identified and segmented per frame, and the process was carried out with high reliability. For instance, there was an inter-annotator agreement rate of $89.2\%$ on gesture identification \cite{rasenberg2022primacy}. This makes this dataset a perfect choice for our study since it allows us to exploit the stroke annotations to devise sequence labels, as explained in Section \ref{sect:preprocessing}. 
The dataset consists of $6106$ gestural strokes with average and median durations of 0.58 and 0.42 seconds, respectively. 
Additional details on the annotated gesture strokes are provided in Section 1 of the Supplementary Materials, and comprehensive descriptions and detailed studies on the nature of the collected data are provided by Eijk \etal \cite{eijk2022cabb}.

While the entire dataset is not fully public due to privacy reasons and the current data-sharing policies of the European Union, we make available the estimated body poses obtained via MMPose \cite{sengupta2020mm} (more details in  Section~\ref{subsect:input_data}) to enable the reproducibility of our results. %

\subsection{Constructing Multi-Phase Sequential Data}
\label{sect:preprocessing}
Using the videos of each participant, we apply a sliding time window approach to generate sequential data.  %
Each time window comprises $18$ frames, reflecting the average gesture stroke duration of $0.58$ seconds at a frame rate of $29.97$ fps. Each time window is shifted with an offset of $2$ frames.
For example, a time window begins at frame $0$ and ends at frame $18$, the subsequent one begins at frame $2$ and ends at frame $20$, and this pattern continues.
Next, we group these overlapping time windows into non-overlapping sequences. Each sequence in our experimental design contains $40$ time windows.
To illustrate, let us consider the first sequence. This sequence will start with the first time window ($0$-$18$) and include up to the $40^{th}$ time window. Given the $2$-frame shift in our sliding window approach, by the time we get to the $40^{th}$ window, we would have covered $80$ frames ($40$ windows $\times$ $2$ frame shift per window). 

Taking as anchor the stroke annotations and building on cognitive science research regarding gestural phases \cite{kendon2004gesture} illustrated in Figure~\ref{fig:gestural_phases}, we label each time window as follows:

\begin{itemize}
[topsep=1.5pt,itemsep=0.5ex,partopsep=1ex,parsep=0ex]
    
    \item Time windows that overlap with the start of a manually annotated stroke for less than $50\%$ are labeled as $P$. Theoretically, this corresponds to the \textit{\textbf{P}reparation} phase, i.e., the transition from a rest position to the most meaningful part of the gesture.
 
    \item Time windows that overlap more than $50\%$ with a manually annotated stroke are labeled as $S$. They are intended to capture the \textit{\textbf{S}troke} phase, where the semantic content of the gesture is typically expressed.
    
    \item Time windows that overlap with the end of a manually annotated stroke for less than $50\%$ are labeled as $R$. Theoretically, this corresponds to the \textit{\textbf{R}etraction} phase, consisting of movements that guide the hands back to a rest position.
    
    \item  Time windows that do not overlap with a manually annotated stroke are labeled as $N$, representing the \textit{\textbf{N}eutral} phases between gestural units (\ie the rest positions).
    
\end{itemize}

This procedure results in the following set of labels: $C = \{P, S, R, N\}$. Note that using the sliding window approach described above, each time window can only have one label. Given the nature of the dataset (where each video corresponds to a full spontaneous dialogue), this results in 
a highly skewed distribution of labels, %
as shown in Table \ref{table:data_distribution}. The dominant label is \textit{Neutral}, representing time windows that do not intersect with any stroke. This is a predictable outcome given the naturalistic character of the dialogues, 
which contain substantial periods without gestures. 
In section \ref{sect:implementation_and_setup}, we explain how we address this challenge.
In addition, there are more stroke samples than boundary samples due to the sliding window's two-frame shift. Given that the average stroke duration is similar to the sliding window's length, there are instances where an entire time window overlaps with a coded stroke.

\begin{table}[h!]
\centering
\resizebox{0.65\columnwidth}{!}{%
    \begin{tabular}{lrr}
    \toprule
    \textbf{Label} & \textbf{\# Samples} & \textbf{Percentage}\\ 
    \midrule
    $P$: \textit{Preparation} & 19103 & 2.3 \%\\
    $S$: \textit{Stroke} & 51325 & 6.0 \%\\
    $R$: \textit{Retraction} & 19212 & 2.3 \%\\ 
    $N$: \textit{Neutral} & 759000 & 89.4 \%\\
    \bottomrule
    \end{tabular}
}
\caption{Label distribution over the entire dataset.}
\label{table:data_distribution}
\end{table}

\begin{figure}[t!]
    \centering\includegraphics[width=0.7\columnwidth]{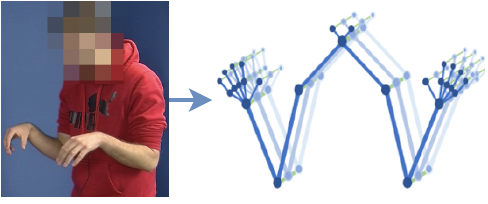}
    \caption{A spatio-temporal graph is extracted from the estimated upper body pose \cite{sengupta2020mm}, adapted from Jiang \etal \cite{jiang2021skeleton}.}
    \label{fig:27_nodes}
\end{figure}

\subsection{Representing Time Windows}
\label{subsect:input_data}
Our method uses upper body skeletal movements from key point estimation generated by applying pose estimation techniques to visual data.
Specifically, we used MMPose to extract body joints from the RGB frames \cite{sengupta2020mm}, which results in the extraction of $133$ body joints. Since our study focuses on manual gesture detection (\ie movements of hands and arms), we only consider the %
upper body key points. %
We follow the approach proposed by Jiang et al. \cite{jiang2021skeleton} for sign language recognition, which involves constructing a Spatio-Temporal (ST)-Graph from 27 upper body joints. ST-Graphs allow us to represent body movement as a sequence of body joints, spatially and temporally, as illustrated in Figure \ref{fig:27_nodes}.
A graph is a data structure consisting of nodes (vertices) and edges connecting the nodes.
In our study, each node carries information about a body joint: the $x$ and $y$ joint positions (in the Cartesian coordinates) and their detection confidence level.
As shown in Figure \ref{fig:27_nodes}, the edges in ST-graphs
are of two types: spatial edges connecting naturally linked joints and temporal edges connecting identical joints across different frames.

\section{Sequence Labeling for Gesture Detection} \label{sect:proposed_model}

As mentioned in previous sections, this study proposes a framework for gesture detection conceptualized as a multi-phase sequence labeling task,  employing the gesture phase labels described in  Section \ref{sect:preprocessing}. Here we define the problem formally, present the architecture of our proposed model, and provide details on baseline models and ablations.
\subsection{Problem Definition}
\label{sec:problem}

Each data point can be denoted as a sequence pair: ($\bm{x}^{(1:t)}, y^{(1:t)}$), where $\bm{x}^{(t)}$ represents the time window located at position $t$ 
and $y^{(t)}$ the corresponding label for that time window. In our study, the input sequence, $\bm{x}^{(1:t)}$, of the body movement time windows is represented as an ST-Graph, and the labels, $y^{(1:t)}$, correspond to the different gesture phases, where $y^{(t)}$ is drawn from the set of discrete labels $C = \{P, S, R, N\}$.

Thus, given an input sequence denoted by $\bm{x}^{(1:t)}$, the sequence labeling's objective is to predict the joint conditional distribution, represented by $\bm{y}^{(1:t)}$. 
As explained in the next sub-section, to leverage the inherent structure of the output sequence, we adopt linear-chain Conditional Random Fields (CRFs), a technique well-suited to exploit the temporal gesture phases in the output sequence %
while predicting the label at a given position $t$. The linearity of this model plays a crucial role, given that we are dealing with sequential data. It conditions only on the preceding transition of the hidden states. 
In other words, the joint conditional distribution of the sequence depends on both the observed states (i.e., the embeddings of the time windows---more details in the next sub-section) and the adjacent predictions of the hidden states (i.e., the phases). Mathematically, this can be represented as:
\begin{align}
    P(\bm{y}^{(1:t)}|{\bm{x}^{(1:t)}}) = \prod_{i=1}^{t} P(\bm{y}^{(i)}|\bm{x}^{(1:t)}, \bm{y}^{(i-1)}).
\end{align}

 \begin{figure}[h]
    \centering
    \scriptsize
    \includegraphics[width=0.9\columnwidth]{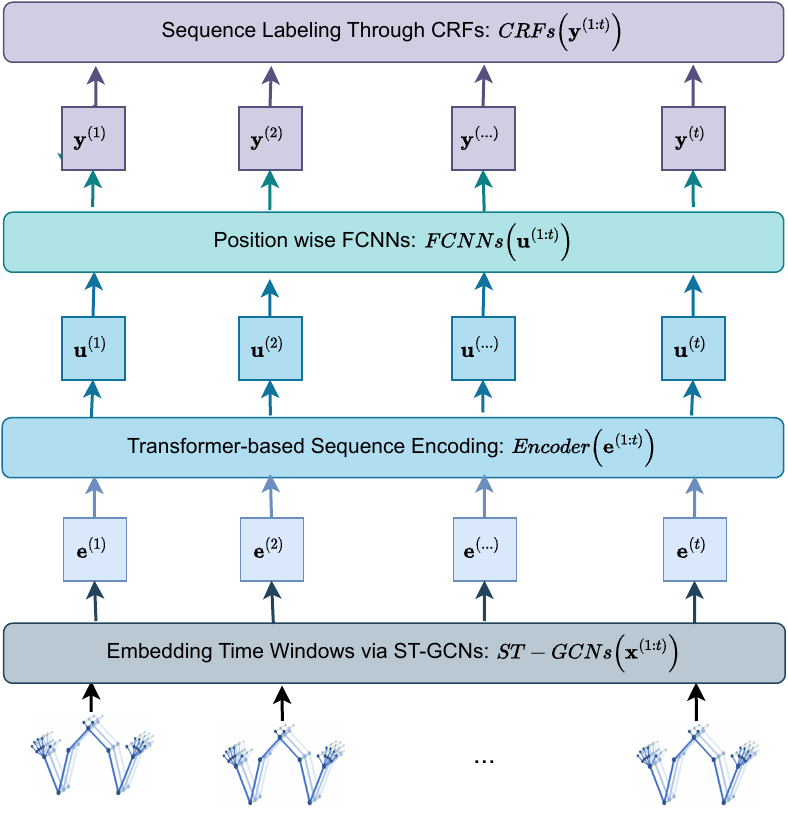}
    \caption{The architecture of the multi-phase sequence labeler consisting of the model 
    components described in Section~\ref{sec:architecture}.} %
    
    \label{fig:main_architecture}
    \vspace{-10pt}
\end{figure}

\subsection{Model Architecture}
\label{sec:architecture}

Our proposed model is illustrated by the diagram in Figure \ref{fig:main_architecture}. %
It employs Spatio-Temporal Graph Convolutional Networks (ST-GCNs) to embed the sequences of time windows. %
A Transformer encoder is then used to learn contextualized embeddings. This is followed by position-wise prediction through fully connected neural networks (FCNNs). Lastly, Conditional Random Fields (CRFs) are applied to leverage sequence labeling on the model's predictions. 

\paragraph{Embedding Time Windows via ST-GCNs}
ST-GCNs, a type of Graph Neural Networks, effectively handle data types such as skeletal key points embedded in graphs, by extending the convolution operation of traditional CNNs to graph-structured data \cite{kipf2017semi, yan2018spatial}. 
Our research uses these networks to process spatio-temporal data represented by skeletal movements in ST graphs.
ST-GCNs capture the movement patterns within the joint sequences through the layered representations of deep neural networks.
They apply the convolution operation on each node in the spatial graph with a c-dimensional vector input feature map, represented by the joint position and detection confidence. 
As depicted in Figure \ref{fig:main_architecture}, this step embeds a sequence of time windows as follows: $\bm{e}^{(1:t)} = GCNs(\bm{x}^{(1:t)})$.
Further details on ST-GCNs, such as their convolutional operations, model layers, and their dimensions, are provided in Section 2 of the supplementary materials. %
A comprehensive description can be found in the seminal work on ST-GCNs by Yan \etal \cite{yan2018spatial}.

\paragraph{Transformer-based Sequence Encoding} 
The Transformer is a neural network architecture with an encoder-decoder structure, known for its state-of-the-art performance in numerous Natural Language Processing and vision tasks \cite{vaswani2017attention}. In our research, we specifically leverage the encoder component of the Transformer, applying it to sequences of body-joint embeddings. The encoder comprises a multi-head self-attention (MHSA) layer followed by position-wise FNNs. 
We use $4$ stacked layers in our architecture. %
Our system operates by feeding sequences of time-window embeddings into the encoder, resulting in contextualized embeddings denoted by $\bm{u}^{(1:t)} = \textit{TransformerEncoder}(\bm{e}^{(1:t)})$. The MHSA layer enhances the model's ability to learn diverse representations at different positions from multiple subspaces.

\paragraph{Position-wise Prediction Layers}
After completing the encoder operations, two additional layers of FCNNs further process each time window's contextual embeddings. The final output layer serves as the prediction layer.
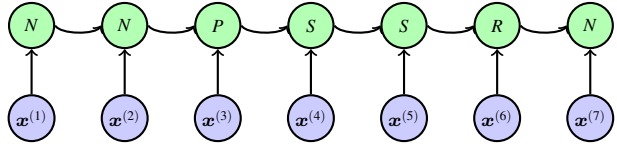
\begin{figure}
\scriptsize
\begin{tikzpicture}[
  node distance=0.6cm,
  state/.style={circle, draw=black, minimum size=0.6cm, fill=green!30, thick, inner sep=0.2em},
  observed/.style={circle, draw=black, minimum size=0.6cm, fill=blue!20, thick, inner sep=0.2em},
  fbox/.style={rectangle, draw=black, inner sep=0.5em, rounded corners, fill=orange!10}
  ]
\node[observed] (X1) {$\bm{x}^{(1)}$};
\node[observed, right=of X1] (X2) {$\bm{x}^{(2)}$};
\node[observed, right=of X2] (X3) {$\bm{x}^{(3)}$};
\node[observed, right=of X3] (X4) {$\bm{x}^{(4)}$};
\node[observed, right=of X4] (X5) {$\bm{x}^{(5)}$};
\node[observed, right=of X5] (X6) {$\bm{x}^{(6)}$};
\node[observed, right=of X6] (X7) {$\bm{x}^{(7)}$};

\node[state, above=of X1] (Y1) {$N$};
\node[state, above=of X2] (Y2) {$N$};
\node[state, above=of X3] (Y3) {$P$};
\node[state, above=of X4] (Y4) {$S$};
\node[state, above=of X5] (Y5) {$S$};
\node[state, above=of X6] (Y6) {$R$};
\node[state, above=of X7] (Y7) {$N$};

\draw[->, thick] (X1) -- (Y1);
\draw[->, thick] (X2) -- (Y2);
\draw[->, thick] (X3) -- (Y3);
\draw[->, thick] (X4) -- (Y4);
\draw[->, thick] (X5) -- (Y5);
\draw[->, thick] (X6) -- (Y6);
\draw[->, thick] (X7) -- (Y7);

\draw[->, thick, bend right=60, looseness=0.6] (Y1.east) to (Y2.west);
\draw[->, thick, bend right=60, looseness=0.6] (Y2.east) to (Y3.west);
\draw[->, thick, bend right=60, looseness=0.6] (Y3.east) to (Y4.west);
\draw[->, thick, bend right=60, looseness=0.6] (Y4.east) to (Y5.west);
\draw[->, thick, bend right=60, looseness=0.6] (Y5.east) to (Y6.west);
\draw[->, thick, bend right=60, looseness=0.6] (Y6.east) to (Y7.west);
\end{tikzpicture}
\caption{Illustration of linear chain CRF for a sequence of 7 states, i.e., gesture phases. Each observed input $\bm{x}^{(i)}$ represents a segment of the video and each state $y^{(i)}$ corresponds to the phase of that segment. The arrows represent the dependencies between the observations and the states and between the states in a sequence.
}
\label{fig:crfs_illustration}
\end{figure}

\paragraph{Structured Prediction via CRFs} 
\label{subsect:crfs}
CRFs are powerful probabilistic models utilized to leverage dependencies between successive labels \cite{lafferty2001conditional}. Figure \ref{fig:crfs_illustration} illustrates the dependencies between gestural phases. For a given sequence label, $\bm{y}^{(1:t)}$, and a set of all possible sequences, $Y_{s}$, the conditional probability of $\bm{y}^{(1:t)}$ is:
\begin{align} 
P(\bm{y}^{(1:t)}|Y_s) = \frac{\exp\left(\sum_{i=1}^{t} f(\bm{y}^{(i-1)}, \bm{y}^{(i)}, s)\right)}{\sum_{\hat{\bm{y}} \in Y_{s}} \exp\left(\sum_{i=1}^{t} f(\hat{\bm{y}}^{(i-1)}, \hat{\bm{y}}^{(i)}, s)\right)},
\label{eq:linear_crf}
\end{align}
where $f(\bm{y}^{(i-1)}, \bm{y}^{(i)}, s)$ represents the transition score from $\bm{y}^{(i-1)}$ to $\bm{y}^{(i)}$, and the score of $\bm{y}^{(i)}$ (\ie, the emissions coming from the prediction layer) for a given sequence $s$. The goal is to optimize the true sequence of labels against the complete set of potential label sequences by maximizing the conditional probability in Equation~\ref{eq:linear_crf}. 
In the prediction step, we solve a combinatorial search problem based on the Viterbi algorithm \cite{forney1973viterbi}, which identifies the sequence path with the highest cumulative probability.

\subsection{Baselines and Ablations}\label{subsect:baselines_and_ablation}

To demonstrate the potential of our multi-phase sequence labeling framework, we compare it against strong baselines. The comparison concerns \textit{two dimensions}: (1) the coarseness of the labeling scheme (multi-phase as in our approach vs.\ binary) and (2) the type of prediction (sequence labeling via CRF vs.\ classification).

In the \textit{binary} approach, we simplify the labeling process by focusing on stroke detection. This strategy aligns with previous research, such as \cite{molchanov2016online, zhu2018continuous, morency2007latent, kopuklu2019real}, where the main interest lies in identifying the gesture stroke in a binary manner.
Time windows that overlap more than $50\%$ with a manually coded stroke are labeled as $S$, while any other time window is labeled as $O$ for \textit{Other}. 
Hence, in a binary model, $y$ is drawn from the set $C = \{S, O\}$.

In the \textit{classification} approach, instead of applying sequence labeling via CRFs, we directly apply a classifier on the embeddings of each time window. Hence, the classifier separately operates on each position's embeddings.
As a result, for predicting the labels of the entire sequence (regardless of whether they are drawn from $\{P, S, R, N\}$ or $\{S, O\}$), we compute the mode of the joint probability distribution, $P(\bm{y}^{(1:t)}|\bm{x}^{(1:t)})$, for each position independently: $ P(\bm{y}^{(1:t)}|{\bm{x}^{(1:t)}})  = \prod_{i=1}^{t} P(\bm{y}^{(i)}|\bm{x}^{(1:l)}).$ The optimization of the classifier can be achieved by minimizing the negative log-likelihood.
In addition to comparing models that differ along these two dimensions, we carry out an ablation on all models to examine the impact of the Transformer encoder.

\section{Experimental Setup}
\label{sect:implementation_and_setup}
This section outlines our solutions to address class imbalance in both the training and testing phases of our framework. Implementation details and models' parameters are presented in Section 2 of the supplementary materials.

\paragraph{Handling Class Imbalance}

To enhance model training while addressing class imbalance (see Table \ref{table:data_distribution}), we employ a sub-sampling approach. Specifically, we use all sequences containing at least one gesture phase and balance them with an equal number of neutral-phase sequences during each training epoch.
We leave the test dataset without changes to check how our model generalizes to imbalanced data.

\paragraph{Evaluation Metrics}
We utilize metrics such as precision, recall, and F1-score designed explicitly to tackle the class imbalance challenge. These metrics are calculated per class, focusing mainly on our classes of interest, such as gesture stroke.
Besides, we use \textit{Intersection over Union (IoU}), which is commonly used in object detection tasks and is also suitable for sequence labeling.

\paragraph{Evaluation Protocol} 
Our evaluation utilizes \textit{subject disjoint} k-fold cross-validation to ensure different subjects (i.e., speakers in a dialogue) across folds. 
That is, the dataset is partitioned into five distinct folds and within each fold, samples from a particular subject are strictly allocated to either the testing or training set.
This division allows us to check the model's generalizability to unseen subjects in the test folds. 
The performance metrics reported in the results and analysis sections are the average scores computed over all five test folds.

\section{Results}
\label{sect:results}
\newcommand{\writeres}[2]{{#1} $\pm$ {#2}}

This section presents our main results. We first focus on performance regarding the detection of the stroke phase and then report the results obtained when detecting gesture units (preparation, stroke, and retraction phases combined).

\begin{table}[t]
\centering
\resizebox{\columnwidth}{!}{%
\begin{tabular}{@{}l@{\hspace{-6pt}}cccccc@{}}
\toprule
\textbf{Approach} & \textbf{Multi-phase} & \textbf{TE} & \textbf{IoU} & \textbf{F1} & \textbf{Precision} & \textbf{Recall} \\
\midrule
\multirow{4}{*}{Classification} & \xmark & \xmark &
\writeres{25.3}{3.8} &
\writeres{40.3}{5.0} &
\writeres{58.7}{2.6} &
\writeres{30.9}{5.1} \\
 & \xmark & \cmark &
\writeres{31.8}{2.7} &
\writeres{48.2}{3.1} &
\writeres{53.1}{2.6} &
\writeres{44.3}{4.2} \\
  & \cmark  &  \xmark &
\writeres{36.0}{2.3} &
52.9 $\pm$ 2.4 & 
46.3 $\pm$ 3.2 & 
\underline{62.0} $\pm$ 2.7 \\
  & \cmark & \cmark &
\writeres{35.4}{1.4} &
52.3 $\pm$ 1.5 & 
41.8 $\pm$ 2.6 & 
\textbf{70.3} $\pm$ 2.4 \\
\midrule
\multirow{4}{*}{\begin{tabular}{@{}c@{}}Sequence\\ labeling\end{tabular}} & \xmark & \xmark & 
\writeres{36.6}{2.4} &
\writeres{53.6}{2.7} &
\writeres{\underline{63.3}}{1.2} &
\writeres{46.5}{3.4} \\
  & \xmark & \cmark &
\writeres{36.2}{1.9} &
\writeres{53.1}{2.0} &
\writeres{\textbf{64.1}}{2.7} &
\writeres{45.6}{3.7} \\
  & \cmark & \xmark &
\writeres{\underline{38.8}}{2.2} &
\underline{55.9} $\pm$ 2.3 & 
59.6 $\pm$ 3.6 & 
53.0 $\pm$ 4.5  \\
\rowcolor{blue!10} & \cmark & \cmark &
\writeres{\textbf{41.2}}{2.0} &
\textbf{58.4} $\pm$ 1.6 & 
55.7 $\pm$ 3.1 & 
61.6 $\pm$ 2.4  \\
\bottomrule
\end{tabular}%
}
\caption{Stroke detection results across all model variants. \textbf{Multi-phase} indicates whether the labeling scheme is multi-phase or binary, and \textbf{TE} whether the Transformer encoder is present or not. The last row corresponds to our proposed framework. The best result for each column in \textbf{bold}, second best \underline{underlined}.}
\label{table:stroke_detection}
\end{table}

\subsection{Gesture Stroke Detection}
An overview of the results on gesture stroke detection is given in Table~\ref{table:stroke_detection}.
The first thing to notice is that all the sequence labeling model variants outperform the classification variants on F1, precision, and IoU scores. Secondly, within each approach (classification vs.\ sequence labeling), all the multi-phase variants outperform the binary variants on F1 and IoU scores. 
Finally, our proposed multi-phase sequence labeling model (\textit{last row of the table}) outperforms all other approaches, with an F1 score of 58.4 and an IoU score of 41.2.
This shows that the two key components of our framework, contextualized sequential prediction and the use of a theoretically motivated multi-phase labeling scheme, independently contribute to enhancing gesture stroke detection.

Regarding the ablation of the Transformer encoder (column TE with \textit{\xmark} in Table~\ref{table:stroke_detection}), we observe 
mixed results. In general, Transformer encoding improves the recall and IoU scores of the models (except for one case, the binary sequence labeler). It can also be seen that removing the Transformer tends to increase 
standard deviations, implying that the model's performance is less consistent across different folds (hence, subjects). This might be due to the Transformer's ability to pay selective attention to different parts of the sequence, allowing it to generalize more effectively to varied data instances. The role of the Transformer will be explained in more detail in the following section.

\begin{table}[h]
\centering
\resizebox{\columnwidth}{!}{%
\begin{tabular}{@{}lccccc@{}}
\toprule
\textbf{Approach} & \textbf{TE} & \textbf{IoU} & \textbf{F1} & \textbf{Precision} & \textbf{Recall} \\
\midrule
\multirow{2}{*}{Classification} & \xmark &
\writeres{46.2}{2.6} &
\writeres{63.2}{2.4} & 
\writeres{60.8}{3.9} & 
\writeres{\underline{66.0}}{2.8} \\
 & \cmark &
\writeres{\underline{48.9}}{1.1} &
\writeres{\underline{65.7}}{1.0} & 
\writeres{57.4}{1.9} & 
\writeres{\textbf{76.8}}{1.6} \\
\midrule

\multirow{2}{*}{\begin{tabular}{@{}c@{}}Sequence labeling \\ labeling\end{tabular}}  & \xmark &

\writeres{43.2}{3.6} &
\writeres{60.2}{3.6} & 
\writeres{\textbf{73.5}}{4.3} & 
\writeres{51.4}{5.2} \\
\rowcolor{blue!10} & \cmark &
\writeres{\textbf{50.6}}{2.1} &
\writeres{\textbf{67.2}}{1.9} & 
\writeres{\underline{72.2}}{3.9} & 
\writeres{63.0}{2.6} \\
\bottomrule
\end{tabular}
}
\caption{
Gesture unit detection results across multi-phase model variants. \textbf{TE} indicates whether the Transformer encoder is present.
The last row corresponds to our proposed framework.
The best result for each metric in \textbf{bold}, second best \underline{underlined}. 
}
\label{table:binary_classification}
\end{table}

\subsection{Gesture Unit Detection}
\label{subsect:gud}

In this subsection, we focus on detecting full gesture units, 
approximated by the combination of the %
preparation, stroke, and retraction phases. 
We analyse the performance of our multi-phase sequential labeling model and its classification counterpart on distinguishing between gesture units as a whole and neutral segments.

Table \ref{table:binary_classification} shows the gesture unit detection results. Our multi-phase sequential labeling approach achieves the best F1 score and IoU, 67.2 and 50.6, respectively. However, regarding recall and precision, two findings emerge. Firstly, using the Transformer encoder noticeably enhances the recall rates across both classification and sequence labeling approaches. This is evidenced by the diminished recall rates observed in the ablated models that exclude the Transformer. Secondly, sequence labeling improves the precision of gesture unit detection.
Interestingly, while the multi-phase sequence labeling model without the Transformer encoder outperforms all classification variants on F1 and IoU scores in the stroke detection task (see Table \ref{table:stroke_detection}), for the task of gesture unit detection its performance is lower than the classification models: 60.2 vs.\ 63.2 and 65.7 F1 scores in Table~\ref{table:binary_classification}.
This may be %
due to its inefficiency in identifying the boundaries of strokes: the preparation and retraction phases. The following section analyzes the reasons behind these patterns, evaluating how these models perform across different gestural phases.

\section{Analysis}
\label{sect:evaluations}
In this section, we zoom into the performance of multi-phase models by analyzing their predictions regarding the internal structure of gestural units. 
\subsection{Performance on each Gesture Phase}

Table \ref{table:class_results} shows the results of the multi-phase models broken down per gesture unit phase (the stroke results are identical to those in Table~\ref{table:stroke_detection}). We clearly see that our multi-phase sequential labeler with the Transformer encoder yields the best performance for the stroke and retraction phases and the second-best results for the preparation phase (where the best model is the ablated version without the Transformer).
However, these detailed results also reveal that the models have more trouble identifying the boundary phases (preparation and retraction) than the stroke: 58.4 F1 score for the stroke vs.\ 35.8 and 28.3 F1 scores for preparation and retraction, respectively. 
The multi-phase sequential model without the Transformer encoder yields particularly low results for the retraction phase: merely 13.4 F1 score. 
The performance of this model on the retraction phase can be explained by its dependency on CRFs. In this model, CRFs depend on previous hidden states (\ie labels) and only on the embeddings of observed states without incorporating context or positional data like order or time (\ie a context provided through the mechanisms of the Transformer Encoder that this model does not employ). We hypothesize that this property affects later hidden states more significantly than earlier ones because of the inherent long sequential dependencies.

The multi-phase classifiers show promising results, particularly for boundary phases, though their performance lags behind in the stroke phase, as already mentioned in Section~\ref{sect:results}. In addition, given that the preparation and retraction phases have a similar number of samples (as shown in Table \ref{table:data_distribution}), comparable classification results are to be expected for both. However, labeling the preparation phase accurately turns out to be simpler than the retraction phase. This consistency was observed across all multi-phase models. Similarly to the sequence labeling models, the classification models also yield lower results for the boundary phases than for the stroke.
A potential reason could be the intricate dependencies of features close to gestures' preparation and retraction phases, which we explore in the following subsection.

\begin{table}[h]
\centering
\resizebox{.9\columnwidth}{!}{%
\begin{tabular}{@{}lcccc@{}}
\toprule
\textbf{Approach} & \textbf{TE} & 
\textbf{Preparation} &
\textbf{Stroke} &
\textbf{Retraction} \\
\midrule
\multirow{2}{*}{Classification} & \xmark &
34.5 $\pm$ 2.5 &
52.9 $\pm$ 2.4 &
20.3 $\pm$ 0.6 \\
 & \cmark &
29.6 $\pm$ 2.3 &
52.3 $\pm$ 1.5 &
\underline{24.9} $\pm$ 3.3 \\
\midrule
\multirow{2}{*}{Sequence labeling} & \xmark &
\textbf{35.8} $\pm$ 1.0 &
\underline{55.9} $\pm$ 2.3 &
13.4 $\pm$ 3.2 \\
&  \cmark &
\underline{34.9} $\pm$ 2.1 &
\textbf{58.4} $\pm$ 1.6 &
\textbf{28.3} $\pm$ 1.6 \\
\bottomrule
\end{tabular}
}
\caption{F1 scores for gesture phases across multi-phase model variants. 
\textbf{TE} indicates whether the Transformer encoder is present or not.
The best result per column in \textbf{bold}, second best \underline{underlined}.}
\label{table:class_results}
\end{table}

\begin{figure}
    \centering
     \includegraphics[width=0.6\columnwidth]{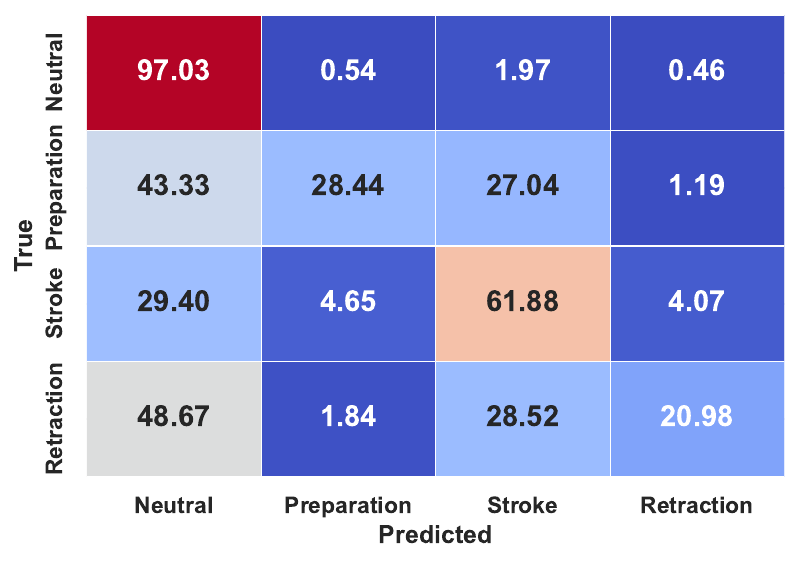}
    \caption{Confusion matrix for the multi-phase sequential labeler.}
    \label{fig:confusion_matrix}
\end{figure}

\subsection{Interaction Between Gesture Phases}
\paragraph{Confusion Matrix} 
Here, we investigate the relationship between different gesture phases using the top-performing model, our proposed multi-phase sequential labeler.
Figure \ref{fig:confusion_matrix} shows that the model predicts the stroke phase very well ($61\%$ accuracy), as already seen in Table \ref{table:stroke_detection}. In addition, the model also predicts the neutral class very accurately ($97\%$); we believe that this is due to the homogeneity of these segments, while the stroke segments are very heterogeneous (including different types of co-speech gestures).
In the previous subsection, we highlighted that the models perform less efficiently in detecting boundary phases than the stroke phase. This observation is also supported by the confusion matrix, which shows that these classes do tend to be confused with neutral and stroke. However, as Table \ref{table:stroke_detection} shows, including them in our approach (multi-phase setting) yields higher stroke detection results, the primary focus of the present study.
Moreover, earlier binary approaches to silent gestures suggested that computational models often confuse the segments around the strokes \cite{molchanov2016online}. Interestingly, our models do not confuse the preparation and retraction phases with each other. The reason could be that these two phases involve different types of movements, \eg, the preparation phase primarily consists of upward hand movements, while the retraction phase involves downward trends. 
To gain more insight into this observation, we next analyze the relationship between gesture phases in the latent space learned by the model.

\begin{figure}[b]
    \centering
    \scriptsize
    \includegraphics[width=0.7\columnwidth]{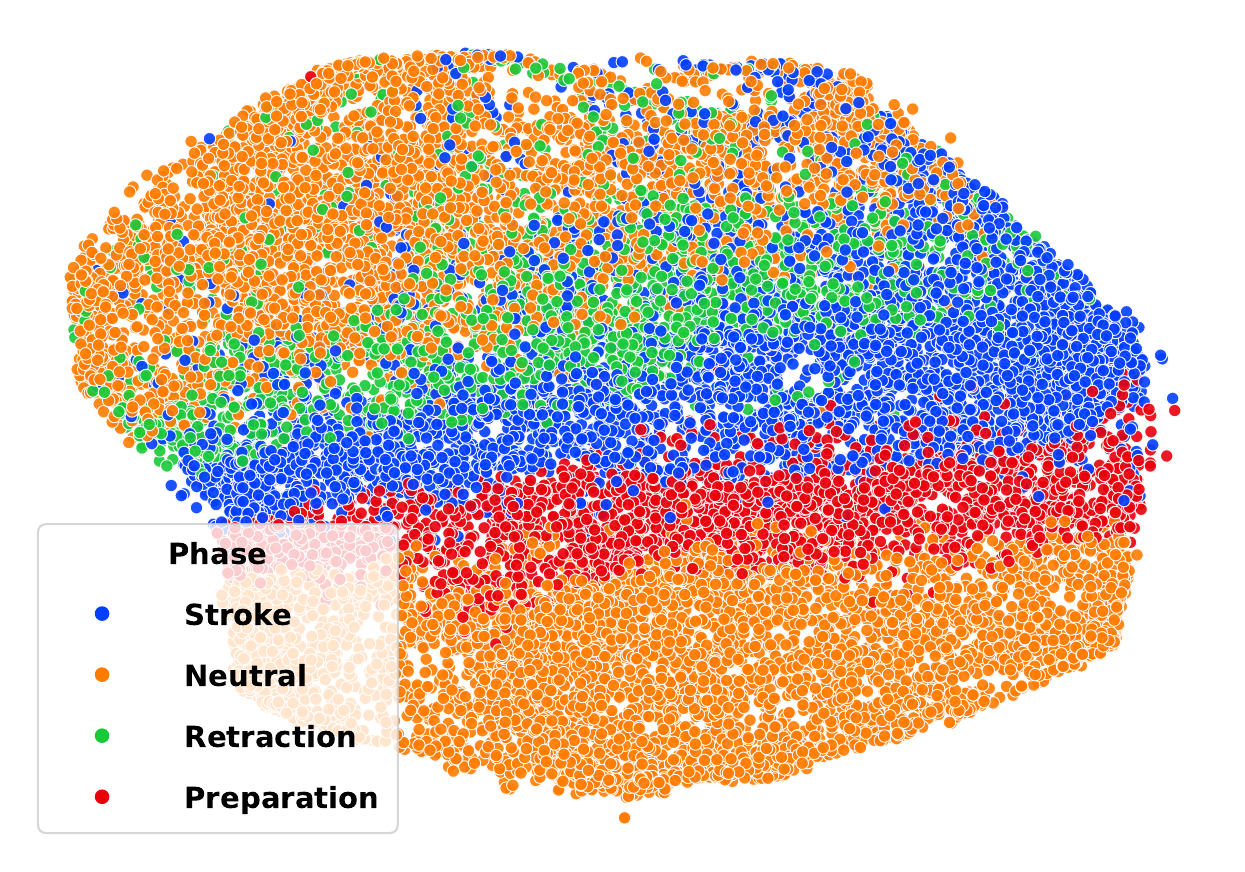}
    \caption{Visualization of gesture phases' embeddings.}    \label{fig:embeddings_visualization}
\end{figure}

\paragraph{Latent-space of Gesture Phases}\label{subsect:tsne}
Figure~\ref{fig:embeddings_visualization} shows the embeddings of sequences of gesture phases 
obtained via t-SNE \cite{van2008visualizing} to visualize the model's latent space. Here, we visualize sequences centered around the stroke, with all phases present. 
The sequences start with a neutral phase, continue with the preparation, stroke, and retraction phases of a gestural unit, and again end with a neutral phase.
The visualization reveals that the %
clusters corresponding to each phase are distinguishable in latent space to varying degrees. 
Notably, the retraction phase, which is characteristically associated with poorer performance across all models compared to other phases, has a less distinct cluster. A closer look %
reveals that this phase's data points mainly relate to neutral sequences and, to a lesser extent, to the stroke phase, a pattern also observed in the confusion matrix in  Figure \ref{fig:confusion_matrix}.

\section{Conclusions}
In this study, we have tackled the problem of automatic gesture detection. Most recent approaches treat this task as binary classification, using data restricted to a limited range of silent gestures. In contrast, we have proposed a novel framework that emphasizes the structured and sequential nature of gestures, focusing on co-speech gestures in naturalistic, conversational data. Our framework reframes the gesture detection task as a multi-phased sequence labeling problem, exploiting a theoretically motivated labeling scheme that distinguishes between the preparation, stroke, and retraction phases of gesture units. 

Our model processes sequences of skeletal movements represented as spatio-temporal graphs over time windows,  uses Transformer encoders to learn contextual embeddings, and leverages Conditional Random Fields to 
perform sequence labeling. In our experiments, we compare our proposed model to strong baselines. 
The results show that sequence labeling methods outperform classification approaches on gesture stroke detection and that the presence of the Transformer encoder improves gesture unit detection for all models, but especially for sequential prediction variants. Finally, we observed that all models are better at detecting the gesture stroke than its boundaries, particularly the retraction phase---an aspect which we aim to improve upon in future work. Overall, our results and analyses highlight our framework's capacity to capture the fine-grained dynamics of co-speech gesture phases, paving the way for more nuanced and accurate automatic gesture detection.

\subsection*{Acknowledgments}
This work was funded by the NWO through a gravitation grant (024.001.006) to the LiI Consortium.
Further funding was provided by the DFG (project number 468466485) and the Arts and Humanities Research Council (grant reference AH/W010720/1) to Peter Uhrig and Anna Wilson (University of Oxford). Raquel Fern\'andez is supported by the European Research Council (ERC CoG grant agreement 819455).
We thank the Dialogue Modelling Group members at UvA, especially Alberto Testoni and Ece Takmaz, for their valuable feedback. We extend our gratitude to Kristel de Laat for contributing to the segmentation of co-speech gestures.
HPC resources are provided by NHR@FAU; see GitHub for details.
{\small
\bibliographystyle{ieee_fullname}
\bibliography{paper}
}
\end{document}


\title{Co-Speech Gesture Detection through Multi-phase Sequence Labeling\\
Supplementary Materials}



\author{
Esam Ghaleb$^{1}$ \; Ilya Burenko$^{2,3}$\; Marlou Rasenberg$^{4,6}$\; Wim Pouw$^{5}$\; Peter Uhrig$^{2,3}$\;\\
Judith Holler$^{5,6}$\; Ivan Toni$^{5}$\; Asl\i~Özy\"{u}rek$^{5,6}$ and Raquel Fern\'{a}ndez$^{1}$\\
$^{1}$University of Amsterdam\;
$^{2}$ScaDS.AI Dresden/Leipzig\; 
$^{3}$TU Dresden\; 
$^{4}$Meertens Institute\\
$^{5}$Radboud University\; 
$^{6}$Max Planck Institute for Psycholinguistics\\
{\tt\small e.ghaleb@uva.nl \;\;raquel.fernandez@uva.nl}
}
\maketitle
\thispagestyle{empty}
\section{Dataset - Gesture Coding and Statistics}

This section provides an overview of the gesture annotation process and the results of the inter-rater reliability check (for gesture identification and coding). The annotation was first leveraged by a study in Rasenberg \etal \cite{rasenberg2022primacy} to study how cross-speaker gestural and lexical alignment emerges.

\paragraph{Gesture Annotation}
In this dataset, only the stroke phase of co-speech gestures was annotated, which is the most meaning-bearing phase \cite{kendon2004gesture, mcneill1992hand}. Gestures were then divided into three categories: (1) iconic gestures that represent physical attributes or actions connected to a referent, (2) deictic gestures, which are often known as pointing gestures, and (3) other gestures, denoting predominantly beat and interactive gestures.

\paragraph{Statistics} The fact that the dataset is collected in a referential game context made iconic gestures the most common category. In detail, the classification of annotated gestures is distributed as follows: a significant majority of 4952 iconic gestures, 360 of other types, and a relatively small count of 145 deictic gestures. Additionally, to distinguish between actual gestures and non-gestures, 642 movement segments were coded, encompassing self-adjustments or hand movements, as non-gestures.
Regarding the length of the annotated strokes, the data indicates an average time of 0.58 seconds, with the most common duration being 0.24 seconds and a median value of 0.41 seconds.

\paragraph{Inter-rater Reliability}
The co-speech gesture coding procedures were conducted in two parts.
The codings were assessed for inter-rater reliability based on trials from the total dataset and involved two independent coders.
In the first part, 96 trials were coded, yielding 296 gesture annotations for comparison. The coders agreed on gesture identification $89.2\%$ of the time. A specialized Staccato algorithm \cite{lucking2011assessing} was used to account for variation in handedness, annotation length, and the number of segments, resulting in scores between 0.71 and 0.80 (on a scale from -1 to 1), suggesting similar segmentation understanding. The second part had a similar procedure, generating 406 gesture comparisons. The inter-rater agreement was slightly lower at $84.7\%$. The same Staccato algorithm was used to standardize segmentation, with the scores ranging between 0.61 and 0.77. 
\section{Implementation \& Model Parameters}\label{sect:sub_model}
 \begin{figure}
    \centering
    \scriptsize
    \includegraphics[width=\columnwidth]{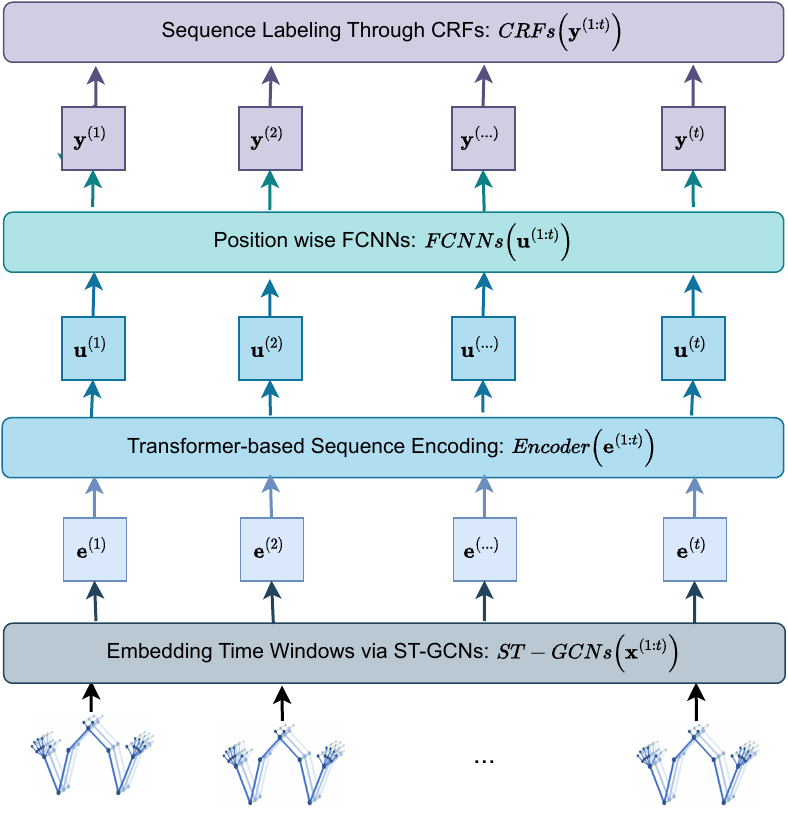}
    \caption{The architecture of the proposed model as outlined in Section \ref{sect:sub_model}.}
    \label{fig:main_architecture}
    \vspace{-15pt}
\end{figure}

\begin{table*}
\centering
\begin{tabular}{lll}
\toprule
\textbf{Layer Type} & \textbf{Layer Details} & \textbf{Dimensions} \\
\midrule
\multirow{3}{*}{ST-GCNs Model} & Input ST-graph & $3\times27\times18$ \\
& Model's 10 layers & Table \ref{table:stgcns}\\
& Output & $256$ \\
\midrule
\multirow{5}{*}{Transformer Encoder} & \textit{Positional encoding} layer input & 256 \\
 & \textit{Positional encoding} sequence max length & 40 \\
 & Four stacked \textit{Transformer Encoders} input & 256 \\
 & \textit{Transformer Encoders} Feed-forward networks & $128\times4$ \\
 & \textit{Transformer Encoders} Attention heads & 8 \\
 & Output & 256 \\
\midrule
\multirow{4}{*}{FCNNs} & Layer 1 Input & 256 \\
 & Layer 1 Output & 128 \\
 & Layer 3 Input & 128 \\
 & Layer 3 Output & 128 \\
 & Layer 5 Input & 128 \\
 & Layer 5 Output & Number of labels \\
\bottomrule
\end{tabular}
\caption{Layer-wise structure and dimensions of the proposed model's components: ST-GCNs, Transformer Encoders, and the Fully Connected Neural Networks.}
\label{table:architecture_parameters}
\end{table*}

\paragraph{Implementation Details}
\label{subsect:implementation_details}
We train all our models with the same set of hyperparameters, which were identified through a process of random search.
We use stochastic gradient descent with 0.1 base learning rate and an $L2$-regularization term with weight $10^{-4}$ to update models' weights. 
We increase the learning rate linearly for the first 20 epochs and divide it by 10 at the $50^{th}$ epoch.
We train models for 80 epochs using 8 Nvidia A100 video cards with a batch size of 256, which was enough for the models to converge.

\paragraph{Model Layers and Dimensions}
The proposed model, as depicted in Figure \ref{fig:main_architecture}, consists of the following components: (1) Spatio-Temporal Graph Convolutional Networks (ST-GCNs), (2) Transformer Encoders, (3) Position Wise Fully Connected Neural Networks (FCNNs), and (4) sequence labeler layer that employs Conditional Random Fields (CRFs). 
Table \ref{table:architecture_parameters} lists the architecture's layers and their dimensions.
Section \ref{subsect:gcns} gives a brief overview of ST-GCNs, and Table \ref{table:stgcns} list our ST-GCNs layers and their dimensions. The full implementation of the model is available in the GitHub repository: {\url{https://github.com/EsamGhaleb/Multi-Phase-Gesture-Detection}}
\begin{table}
\centering
\begin{tabular}{|c|c|c|c|}
\hline
\textbf{Layer} & \textbf{In Channel} & \textbf{Out Channel} & \textbf{Stride} \\
\hline
l1 & 3 & 64 & None \\
\hline
l2 & 64 & 64 & None \\
\hline
l3 & 64 & 64 & None \\
\hline
l4 & 64 & 64 & None \\
\hline
l5 & 64 & 128 & 2 \\
\hline
l6 & 128 & 128 & None \\
\hline
l7 & 128 & 128 & None \\
\hline
l8 & 128 & 256 & 2 \\
\hline
l9 & 256 & 256 & None \\
\hline
l10 & 256 & 256 & None \\
\hline
\end{tabular}
\caption{ST-GCNs layers parameters and dimensions \cite{jiang2021skeleton}.}
\label{table:stgcns}
\end{table}

\paragraph{ST-GCNs}\label{subsect:gcns}
GCNs, a subset of graph neural networks, are models that have emerged from the success of traditional Convolutional Neural Networks (CNNs). They extend the convolution operation (template matching) from CNNs to accommodate data structured as graphs, allowing them to handle data with varying structures. 
GCNs are ideal for data structures that can be represented as graphs, such as spatio-temporal graphs of body joints, social networks or molecular structures \cite{yan2018spatial}.
In our research, GCNs prove particularly useful given the nature of our data, which uses ST-graphs to represent skeletal movements. 

Technically, ST-GCNs extend conventional convolution operations to GNNs with features represented on a spatial graph $V$. The input feature map $f_{in}$ at frame $t$ is a c-dimensional vector for each node in the graph. For instance, in our study, $c$ is a three-dimensional vector of each joint position (\ie $x$ and $y$) and the joint detection confidence at the input layer.
The convolution operation is performed for each node $v_{i}$ according to the formula: $f_{out}(v_i) = \sum_{v_j \in B_i} \frac{1}{Z_{ij}} f_{in}(v_j) .w(l_i(v_j))$ where $B_i$ represents the neighboring nodes $v_j$, $Z_{ij}$ is a normalization term, $w$ is a learnable kernel, and $l_i$ maps the weight vectors for each vertex. 

In the spatial context, ST-GCNs adopt spatial configuration partitioning to map weights, creating three subsets, denoted as $K$: the root node, a centripetal group, and centrifugal nodes.
The convolution operation, in its vectorized form, is defined as $\bm{f}_{out} = \sum_k^{K_v} \bm{A}_k \odot \bm{M}k (\bm{f}{in} \bm{W}_k)$ where $K_v$ is the kernel size, $\bm{A}_k$ is the adjacency matrix normalized by $\hat{D}^{-\frac{1}{2}}$, and $\bm{M}$ and $\bm{W}$ are learnable matrices.
Temporal convolution is implemented with an $L \times 1$ convolutional layer to learn features from adjacent frames.

In our implementation, we used a pre-trained model for sign language recognition by Jiang \etal~\cite{jiang2021skeleton}. Table \ref{table:stgcns} lists the ST-GCNs model's layers and dimensions. Finally, a comprehensive description of this model can be found in the seminal work on ST-GCNs by Yan \etal \cite{yan2018spatial}.

{\small
\bibliographystyle{ieee_fullname}
\bibliography{paper}
}